\documentclass[preprint,11pt]{elsarticle}

\usepackage[utf8]{inputenc}
\usepackage[margin=1.5cm]{geometry}
\usepackage{titlesec}
\usepackage{tabu}
\usepackage{enumitem}
\usepackage{amssymb}
\usepackage{xcolor}
\usepackage{pythonhighlight}
\usepackage{xspace}
\usepackage{graphicx}

\newlist{selectlist}{itemize}{2}
\setlist[selectlist]{label=$\square$,leftmargin=*,noitemsep,topsep=0pt}

\usepackage{lmodern}
\usepackage{hyperref,cleveref}
\hypersetup{
    colorlinks=true,
    linkcolor=blue,
    filecolor=magenta,      
    urlcolor=blue,
}
 
\titleformat{\section}[block]{\hspace{1em}\bfseries}{\thesection.}{0.5em}{} 
\titleformat{\subsection}[block]{\hspace{1em}}{\thesubsection}{0.5em}{}

\usepackage[nolist]{acronym}
\begin{acronym}[UML]
	\acro{AI}{Artificial Intelligence}
	\acro{CEL}{Class Expression Learning}
	\acro{CL}{Concept Learning}
	\acro{DL}{Description Logic}
    \acro{KB}{Knowledge Base}
    \acro{KG}{Knowledge Graph}
    \acrodefplural{KG}{Knowledge Graphs}
    \acro{KGE}{Knowledge Graph Embedding}
	\acro{ILP}{Inductive Logic Programming}
	\acro{RL}{Reinforcement Learning}
	\acro{OWL}{Web Ontology Language}
	\acro{MDP}{Markov Decision Process}
	\acro{CWA}{Closed World Assumption}
	\acro{ML}{Machine Learning}
\end{acronym}  

\newcommand{\kg}{\ensuremath{\mathcal{G}}\xspace}
\newcommand{\emb}[1]{\ensuremath{\mathbf{#1}}}
\newcommand{\triple}[3]{(\texttt{#1}, \texttt{#2}, \texttt{#3})}
\newcommand{\entities}{\ensuremath{\mathcal{E}}\xspace}
\newcommand{\relations}{\ensuremath{\mathcal{R}}\xspace}
\newcommand{\scoreFunc}{\phi}

\begin{document}
\noindent

\noindent
\textbf{Hardware-agnostic Computation for Large-scale Knowledge Graph Embeddings}
\vskip0.5cm
\noindent
\textbf{Caglar Demir\footnote{Corresponding Author} and Axel-Cyrille Ngonga Ngomo}\\
Data Science Group, Paderborn University, Germany\\

\noindent
\textbf{Abstract}\\
Knowledge graph embedding research has mainly focused on learning continuous representations of knowledge graphs towards the link prediction problem.
Recently developed frameworks can be effectively applied in research related applications.
Yet, these frameworks do not fulfill many requirements of real-world applications.
As the size of the knowledge graph grows, moving computation from a commodity computer to a cluster of computers in these frameworks becomes more challenging.
Finding suitable hyperparameter settings w.r.t. time and computational budgets are left to practitioners.
In addition, the continual learning aspect in knowledge graph embedding frameworks is often ignored, although continual learning plays an important role in many real-world (deep) learning-driven applications.
Arguably, these limitations explain the lack of publicly available knowledge graph embedding models for large knowledge graphs.
We developed a framework based on the frameworks DASK, Pytorch Lightning and Hugging Face to compute embeddings for large-scale knowledge graphs in a hardware-agnostic manner, which is able to address real-world challenges pertaining to the scale of real application.
We provide an open-source version of our framework along with a hub of pre-trained models having more than 11.4 B parameters\footnote{\url{https://github.com/dice-group/dice-embeddings}}.
\noindent
\textbf{Code metadata}\\
\noindent
\begin{tabular}{|l|p{6.5cm}|p{9.5cm}|}
\hline
\textbf{Nr.} & \textbf{Code metadata description} & \textbf{Please fill in this column} \\
\hline
C1 & Current code version & v3 \\
\hline
C2 & Permanent link to code/repository used for this code version & \url{https://github.com/dice-group/dice-embeddings}\\
\hline
C3  & Permanent link to Reproducible Capsule & \url{https://codeocean.com/capsule/6862303/tree/v1}\\
\hline
C4 & Legal Code License   & AGPL-3.0 license \\
\hline
C5 & Code versioning system used & git, gitflow \\
\hline
C6 & Software code languages, tools, and services used & Python, Pytorch, Pytorch-Lightning, DASK, Hugging Face, Pandas, Numpy, and more\\
\hline
C7 & Compilation requirements, operating environments \& dependencies & 
\url{https://github.com/dice-group/dice-embeddings#installation}
\\
\hline
C8 & If available Link to developer documentation/manual & 
\url{https://github.com/dice-group/dice-embeddings#documentation}\\
\hline
C9 & Support email for questions & caglar.demir@upb and caglardemir8@gmail.com\\
\hline
\end{tabular}\\
\vskip0.5cm
\noindent
\section{Introduction}
\acp{KG} represent structured collections of facts~\cite{hogan2020knowledge} and are being used in many challenging applications, including web search, question answering, and recommender systems~\cite{nickel2015review}. Despite their usefulness in many applications, most knowledge graphs are incomplete, i.e., contain missing links. The task of identifying missing links in knowledge graphs is referred to as~\textit{link prediction}. 
In the last decade, a plethora of \ac{KGE} approaches have been successfully applied to tackle various tasks including the link prediction task~\cite{nickel2015review}. \ac{KGE} models aim to learn continuous vector representations (embeddings) for entities and relations tailored towards the link prediction task. 


\textbf{Knowledge Graph Embeddings and Challenges:}
Let $\entities$ and $\relations$ represent the sets of entities and relations. A \ac{KG} is often formalised as a set of triples $\kg= \{\triple{h}{r}{t} \}  \subseteq \entities \times \relations \times \entities$ where each triple contains two entities $\texttt{h},\texttt{t} \in \entities$ and a relation $\texttt{r} \in \relations$~\cite{demir2021convolutional,nickel2015review}. 
Most \ac{KGE} models are defined as a parametrized scoring function $\scoreFunc_\Theta: \entities \times \relations \times \entities \rightarrow \mathbb{R}$ such that $\scoreFunc_\Theta\triple{h}{r}{t}$ ideally signals the likelihood of \triple{h}{r}{t} is true~\cite{nickel2015review}.
In a simple setting,
$\Theta$ contains an entity embedding matrix $\mathbf{E} \in \mathbb{R}^{ |\entities| \times d}$ and
a relation embedding matrix $\mathbf{R} \in \mathbb{R}^{ |\relations| \times d}$, where $d$ stands for the embedding vector size.
Given the triples \triple{Barack}{Married}{Michelle} and $\triple{Michelle}{HasChild}{Malia}\in \kg$, a good scoring function is expected to return high scores for \triple{Barack}{HasChild}{Malia} and
\triple{Michelle}{Married}{Barack}, while returning a considerably lower score for \triple{Malia}{HasChild}{Barack}.
To compute a single score, embeddings of entities and relations are retrieved from $\mathbf{E}$, $\mathbf{R}$ and trilinear d-dimension vector multiplication is performed, i.e., $\scoreFunc\triple{Barack}{HasChild}{Malia}=\emb{Barack} \circ \emb{HasChild} \cdot \emb{Malia}$ (see \cite{demir2022kronecker}).

As $|\kg|$ increases, the total training time of learning good representations $\Theta$ increases. This magnifies the importance of effective parallelism. This is often realised as data parallelism which stores a copy of a \ac{KGE} model in available CPUs or GPUs.
Most available frameworks including \ac{KGE} frameworks rely on this paradigm (see PyKEEN~\cite{ali2021pykeen} and libkge~\cite{libkge}. A $|\kg|$, $|\entities|$ and $|\relations|$ grow, $\Theta$ often does not fit in a GPU. This limitation gives a rise to model parallelism and sharded training.
Through FairScale\cite{FairScale2021} or DeepSpeed techniques \cite{rasley2020deepspeed} provided within Pytorch Lightning~\cite{falcon2019pytorch},
our framework effectively partitions $\Theta$ into available CPUs and GPUs, instead of plain cloning. 
This allows to train gigantic models ($>$11 B parameters).
PyKEEN~\cite{ali2021pykeen} and libkge~\cite{libkge} lack of the model parallelism feature among many other features such as the deployment service.

\section{Description}
\label{sec:description}
\textbf{Hardware-agnostic Computation:}
The core goal of our framework is to facilitate learning large-scale knowledge graph embeddings in an hardware-agnostic manner. 
Hence, practitioners can use our framework to learn embeddings of \ac{KG} on commodity hardware as well as a cluster of computers without chaining a single line of code.
We based our framework on Pytorch Lightning~\cite{falcon2019pytorch} and DASK~\cite{dask}.
Pytorch Lightning allows our framework to use multi-CPUs,-GPU and even -TPUs in an hardware-agnostic manner.
This implies that many important decisions at finding a suitable configuration for the learning process can be made automatically, i.e., finding a batch size that optimally fits in to the memory, scaling to cluster of computers.
We observe that most embedding frameworks rely on a single core while reading and preprocessing the input \ac{KG}.
As the size of \ac{KG} grows, this design decision becomes an increasing hindrance to scalability and increases the total runtime.
Moreover, the process of the reading, preprocessing, and indexing an input \ac{KG} is often intransparent to practitioners.
That means that as the size of \ac{KG} grows, practitioners are not informed about the current stage of the total computation.
Through a DASK dashboard, our framework shares details about all steps of computation with users.
The DASK dashboard can be used to used to analyse the reading and preprocessing steps in a fine-grained manner when computation is moved from a commodity computer to cluster of computers. 
\textbf{Finding suitable configuration:}
Our framework dynamically suggests a suitable configuration setting for a given dataset and available computational resources. This includes many features, e.g., finding most memory efficient integer data type for indexing, batch size, embedding vector size as well as learning rate for a given input configuration. Currently, we are working on forging our framework with Auto-Machine Learning techniques to facilitate the usage of framework by novice users. 
By this, we aim to share our expert knowledge with practitioners to that their computational and time budgets can be effectively utilized.
Computational and time budgets of practitioners play an important role in real-world successful \ac{ML} applications~\cite{bottou2007tradeoffs}.

\textbf{Continual Learning and Deployment:}
Our framework continues to assist practitioners after the embedding learning process.
In many \ac{ML} applications, the input data evolves with the time. 
Hence, continual learning plays an important role in successful applications of \ac{ML} models\cite{diethe2019continual}. 
Yet, most \ac{KGE} frameworks do not provide means for continual learning.
To alleviate this issue, our framework can be used to train \ac{KGE} models on non-static data.
Moreover, practitioners can deploy their model in a web-application without writing a single line of code as in our github repostiory.
%

\textbf{Extendability:}
The software design of our framework allows practitioners to solely focus on their novel ideas, instead of engineering.
For instance, a new model can be implemented in our framework without an effort. 
Inheriting from BaseKGE class, a new embedding model can be readily included into our framework. 
\begin{figure}
\centering
\small
\begin{python}
class ComplEx(BaseKGE):
    def __init__(self, args):
        super().__init__(args)
        self.name = 'ComplEx'
    def forward_triples(self, x: torch.Tensor) -> torch.Tensor:
        # (1) Retrieve Embedding Vectors
        head_ent_emb, rel_ent_emb, tail_ent_emb = self.get_triple_representation(x)
        # (2) Split (1) into real and imaginary parts.
        emb_head_real, emb_head_imag = torch.hsplit(head_ent_emb, 2)
        emb_rel_real, emb_rel_imag = torch.hsplit(rel_ent_emb, 2)
        emb_tail_real, emb_tail_imag = torch.hsplit(tail_ent_emb, 2)
        # (3) Compute Hermitian inner product.
        real_real_real = (emb_head_real * emb_rel_real * emb_tail_real).sum(dim=1)
        real_imag_imag = (emb_head_real * emb_rel_imag * emb_tail_imag).sum(dim=1)
        imag_real_imag = (emb_head_imag * emb_rel_real * emb_tail_imag).sum(dim=1)
        imag_imag_real = (emb_head_imag * emb_rel_imag * emb_tail_real).sum(dim=1)
        return real_real_real + real_imag_imag + imag_real_imag - imag_imag_real
\end{python}
    \caption{Including an implementation of state-of-the-art Embedding model in our framework.}
    \label{fig:my_label}
\end{figure}

\textbf{Summary of Initial Experimental Results:}
We used the most recent DBpedia 2021 benchmark dataset\footnote{\url{https://databus.dbpedia.org/dbpedia/collections/dbpedia-snapshot-2021-06}} to evaluate our framework in depth. 
Our experiments suggest that a state-of-the-art \ac{KGE} model with more than 11.4B parameters can be successfully trained and applied in link prediction, and relation prediction
\footnote{\url{https://hobbitdata.informatik.uni-leipzig.de/KGE/DBpediaQMultEmbeddings_03_07/}}. We refer to the project page for the details and log files about pretrained models.

\section{Software Impact}
\label{sec:software_impact}
Our framework facilitates the use of \ac{KGE} models on large \acp{KG} without requiring expert knowledge in software engineering. Hence, it helps practitioners to spend more time on generating value by using embedding models, instead of investing it into engineering for large-scale experiments.
Our framework is now being deployed in real use cases within the funded research projects mentioned in the acknowledgements.

\textbf{Limitations and future improvements:}
We plan to investigate (1) pseudo-labeling to leverage unlabelled data, (2) Auto-ML for the embedding model design, and (3) state-of-the-art continue learning techniques.

\textbf{Scholarly Publications:}
Our framework have been effectively used to learn knowledge graphs embeddings in several published works~\cite{pmlr-v157-demir21a},\cite{demir2021convolutional},\cite{demir2021shallow},\cite{demir2022kronecker},\cite{Zahera2022Tab2Onto},\cite{kouagou2021learning}, and \cite{Heindorf2021EvoLearner}.

\textbf{Acknowledgements:}
This work has been supported by the German Federal Ministry of Education and Research (BMBF) within the project DAIKIRI under the grant no 01IS19085B and by the the German Federal Ministry for Economic Affairs and Energy (BMWi) within the project RAKI under the grant no 01MD19012B.
\noindent

\noindent
\bibstyle{elsarticle-num} 
\bibliography{mybibfile}
\end{document}